\newcommand{\CLASSINPUTtoptextmargin}{.75in}
\newcommand{\CLASSINPUTbottomtextmargin}{.55in}
\newcommand{\CLASSINPUTinnersidemargin}{.68in}
\newcommand{\CLASSINPUToutersidemargin}{0.68in}
\documentclass[conference,9pt]{IEEEtran}
\IEEEoverridecommandlockouts
\usepackage{cite}
\usepackage{amsmath,amssymb,amsfonts}
\usepackage{algorithmic}
\usepackage{graphicx}
\usepackage{textcomp}
\usepackage{xcolor}
\usepackage{multirow}
\usepackage{booktabs}
\usepackage{epstopdf}
\usepackage[caption=true,font=footnotesize]{subfig}
\usepackage{cite}
\usepackage{url}
\newcommand{\STAB}[1]{\begin{tabular}{@{}c@{}}#1\end{tabular}}

\def\BibTeX{{\rm B\kern-.05em{\sc i\kern-.025em b}\kern-.08em
    T\kern-.1667em\lower.7ex\hbox{E}\kern-.125emX}}

\renewcommand{\baselinestretch}{.97}
\begin{document}

\title{Contextual-Bandit Anomaly Detection for IoT Data in Distributed Hierarchical Edge Computing}

\author{
    \IEEEauthorblockN{%
        Mao~V.~Ngo\IEEEauthorrefmark{1}\IEEEauthorrefmark{2},
        Tie~Luo\IEEEauthorrefmark{3},
        Hakima~Chaouchi\IEEEauthorrefmark{4}, and
        Tony~Q.S.~Quek\IEEEauthorrefmark{1}}%
    \IEEEauthorblockA{\IEEEauthorrefmark{1}%
\small Singapore University of Technology and Design,
        Singapore}%
    \IEEEauthorblockA{\IEEEauthorrefmark{2}%
\small Institute for Infocomm Research,
        A*STAR,
        Singapore}%
    \IEEEauthorblockA{\IEEEauthorrefmark{3}%
\small Department of Computer Science,
        Missouri University of Science and Technology,
        USA}%
    \IEEEauthorblockA{\IEEEauthorrefmark{4}%
\small CNRS, SAMOVAR, Telecom Sud Paris, Institut Mines Telecom,
        Paris-Saclay University,
        France}%
    \IEEEauthorblockA{%
\small         \texttt{vanmao\_ngo@mymail.sutd.edu.sg},
        \texttt{tluo@mst.edu},\\
        \texttt{hakima.chaouchi@telecom-sudparis.eu},
        \texttt{tonyquek@sutd.edu.sg}}%
}

\maketitle

\begin{abstract}
Advances in deep neural networks (DNN) greatly bolster real-time detection of anomalous IoT data.
However, IoT devices can hardly afford complex DNN models, 
and offloading anomaly detection tasks to the cloud incurs long delay. 
In this paper, we propose and build a demo for an adaptive anomaly detection approach for distributed hierarchical edge computing (HEC) systems to solve this problem, for both univariate and multivariate IoT data.
First, we construct multiple anomaly detection DNN models with increasing complexity, and associate each model with a layer in HEC from bottom to top. 
Then, we design an adaptive scheme to select one of these models on the fly, based on the contextual information extracted from each input data.
The model selection is formulated as a {\em contextual bandit problem} characterized by a single-step Markov decision process, and is solved using a {\em reinforcement learning policy network}.
We build an HEC testbed, implement our proposed approach, and evaluate it using real IoT datasets. The demo shows that our proposed approach significantly reduces detection delay (e.g., by 71.4\% for univariate dataset) without sacrificing accuracy, as compared to offloading detection tasks to the cloud.
We also compare it with other baseline schemes and demonstrate that it achieves the best accuracy-delay tradeoff. 
\end{abstract}

\section{Introduction}
With the increasing demand of detecting anomalous sensory data generated by a massive number of IoT devices, machine learning---especially deep learning---offers an effective approach and has been successfully applied to many anomaly detection tasks in IoT environments~\cite{Luo_ICC2018,Malhotra_LSTM_encDec_ICMLWrsh2016, Mao2020adaptive}.
A variety of IoT applications, such as collision avoidance for autonomous vehicles and fire alarm systems in factories, are time-critical and require fast anomaly detection.
In these cases, the traditional approach of streaming all the IoT sensory data to the cloud can be problematic as it tends to incur high communication delay, congest backbone networks, and pose data privacy threats.

Anomaly detection (AD) with edge or fog computing \cite{La_FogComputing_2019,Chen_SEC2017} provides an alternative by performing distributed AD in the proximity of sensory data sources.
However, pushing computation from cloud to the edge faces resource challenges especially when complex deep learning models are used. 
Remedies include 
model compression \cite{Han2015DeepCompression}
or successive offloading in a hierarchy until a certain performance threshold is reached \cite{Teerapittayanon_ICDCS2017}.
But overall, there are three main issues in existing works:
(1) ``one size fits all'' - attempting to use one AD model to handle all the input data, while overlooking the fact that different data samples often have different levels of hardness in detecting anomaly events;
(2) focusing on accuracy or F1-score without giving adequate consideration on detection delay and memory footprint;
(3) lacking appropriate local analysis and thus often transmitting data back and forth between sources and the cloud, incurring unnecessary delay and bandwidth consumption.


In this paper, we propose an adaptive distributed approach that leverages the hierarchical edge computing (HEC) architecture by adaptively matching data of different hardness levels of detection with models of different complexity.
Specifically, we construct multiple anomaly detection DNN models (using autoencoder and LSTM) of increasing complexity and associate them with the multiple layers of HEC from bottom to top, i.e., IoT devices, edge servers, and the cloud.
Then, we propose an adaptive scheme that judiciously selects one of these models to perform AD at the most suitable layer, based on the contextual information extracted from each input data on the fly.
The scheme follows a single-step Markov decision process (hence can make quick decisions) derived from a {\em contextual bandit problem} that we formulate and solve using a {\em reinforcement learning policy network}.
By selecting appropriate models, we avoid unnecessary data transmissions between IoT devices, edge servers, and the cloud, while maintaining the best detection accuracy.
We build an HEC testbed and implement our proposed approach for both univariate and multivariate data.\footnote{\label{url_demo}Our demo is also available online: \url{https://rebrand.ly/91a71}} 
Our extensive evaluation using real-world IoT datasets demonstrate that our proposed approach achieves the best tradeoff between high detection accuracy and low detection delay, and outperforms multiple other benchmark schemes.

\section{Design}
\label{sec:AdaptiveAnomalyDetection}

\begin{figure*}[tb]
	\centering
    \subfloat[Testbed hierarchical edge computing (HEC) and architecture of anomaly detection (AD) models]{%
	   \includegraphics[width=0.66\linewidth]{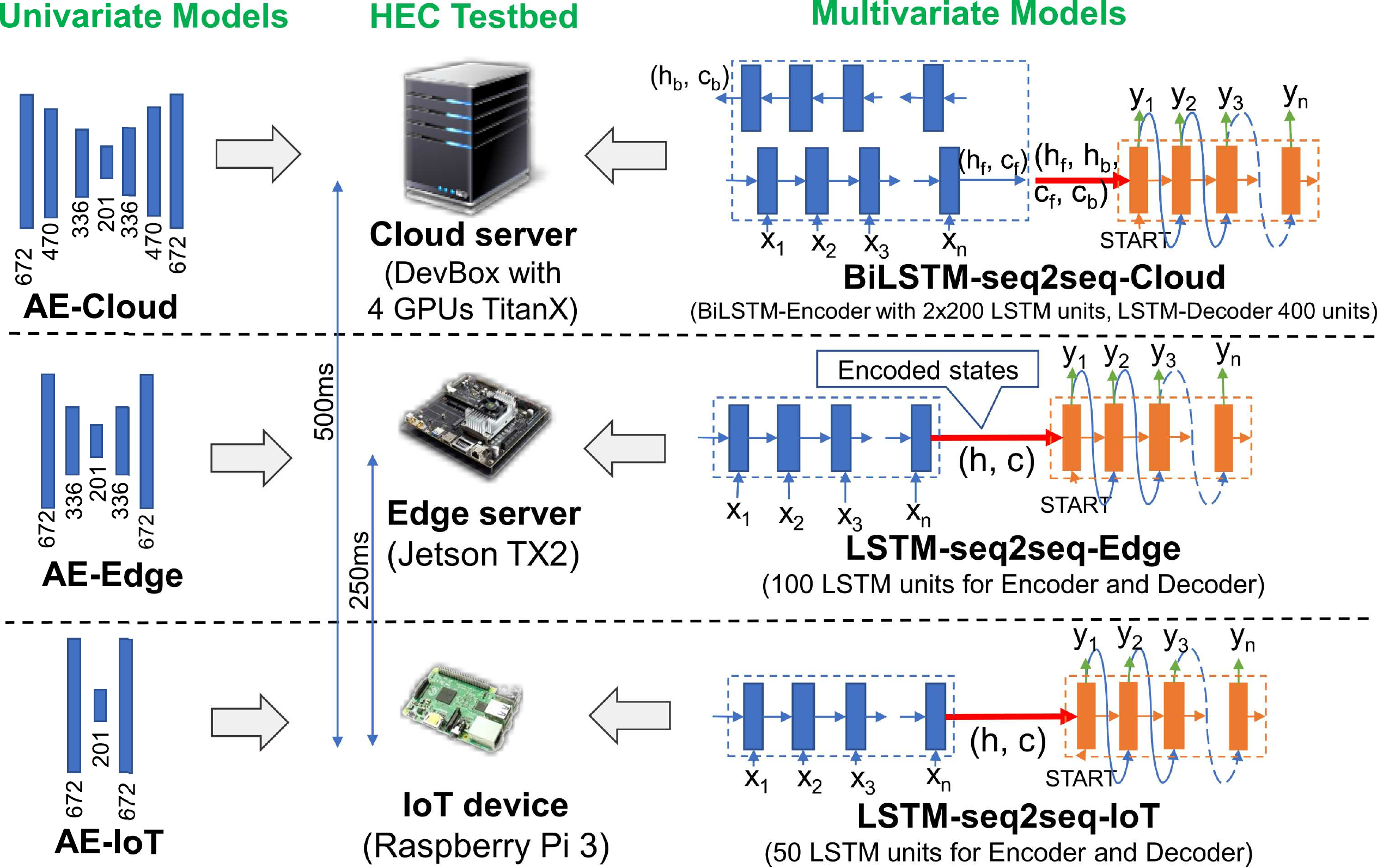}%
	   \label{fig:OverviewSystem}%
    }\hfill
    \subfloat[Software architecture of the demo.]{%
        \includegraphics[width=0.32\linewidth]{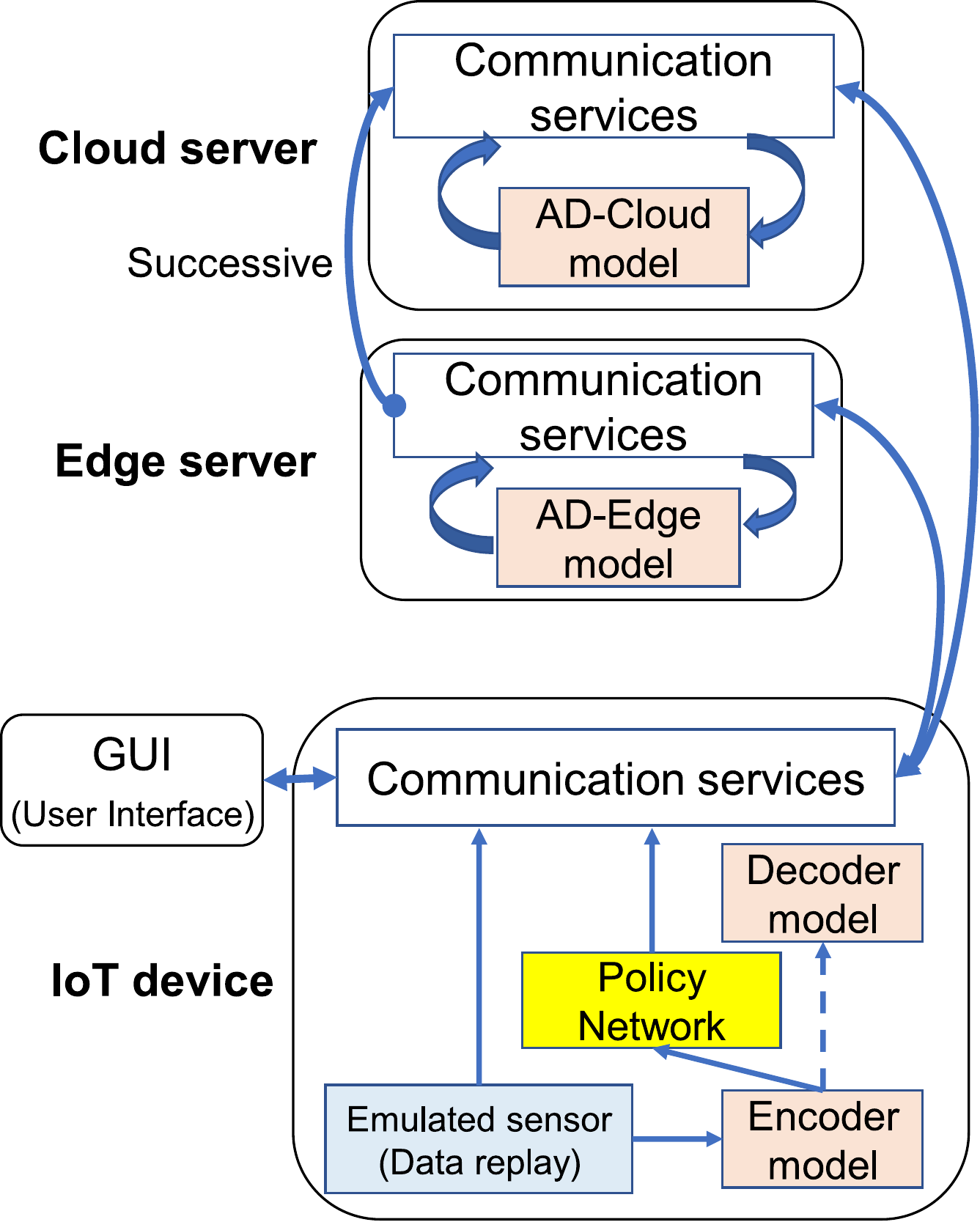}%
        \label{fig:SoftwareArchitecture}
    }
 \caption{The HEC testbed and architecture of AD models for univariate and multivariate data. Demo software architecture.}
\end{figure*}

We consider a $K$-layer distributed hierarchical edge computing (HEC), and choose $K=3$ as a typical setting \cite{La_FogComputing_2019} (but our approach applies to any $K$ in general, i.e., multiple layers of edge servers).
We build an HEC testbed with three layers consisting of a Raspberry Pi 3 as the IoT device, an NVIDIA Jetson-TX2 as the edge server, and an NVIDIA Devbox (with 4 GPU TitanX) as the cloud server, as shown in Fig.~\ref{fig:OverviewSystem}.

We consider two types of data: univariate and multivariate.
We construct $K=3$ AD models for each type, with increasing complexity, and associate those models with the HEC layers from 1 to $K$.
Then, we design an adaptive model selection scheme to choose the most suitable model on the fly to detect anomalies.

\subsection{Constructing Anomaly Detection Models in HEC}
\label{subsec:multipleADModels}

\subsubsection{AD Models for Univariate Data}
We use autoencoders (AE) to build AD models for univariate IoT data, as the feasibility of running such models on IoT devices has been proved by \cite{Luo_ICC2018}. Hence, we build three AE-based models called \textit{AE-IoT}, \textit{AE-Edge}, and \textit{AE-Cloud}, respectively, to associate with the three corresponding layers of HEC. These models have three, five, seven layers and thus have different capabilities of learning features for data representation. See Fig.~\ref{fig:OverviewSystem}.

\subsubsection{AD Models for Multivariate Data}
The simplicity of AE-based models does not well represent features for high-dimensional IoT data. Hence in the
multivariate case (18 dimensions in our demo), we use sequence-to-sequence (seq2seq) model \cite{Sutskever2014SequenceTS} based on long short-term memory (LSTM) to build an LSTM encoder-decoder as the AD model.
Such models can learn representations of multi-sensor time-series data with 1 to 12 dimensions \cite{Malhotra_LSTM_encDec_ICMLWrsh2016}. In our case, we apply our LSTM-seq2seq model to an 18-dimensional dataset, and deploy the model on the IoT device, and name it \textit{LSTM-seq2seq-IoT}.
The multivariate IoT data are encoded into encoded states by an LSTM-encoder, then an LSTM-decoder learns to reconstruct data from previous encoded states and previous output, one step a time (for the first step, a special token is used, which in our case is a zero vector).
For the edge layer, we build an \textit{LSTM-seq2seq-Edge} anomaly detection model with double number of LSTM units for both encoder and decoder, which can learn a better representation of a longer sequence input.
For the cloud layer, we build a \textit{BiLSTM-seq2seq-Cloud} anomaly detection model with a bidirectional-LSTM (BiLSTM) encoder to learn both backward and forward directions of the input sequence to encode information into encoded states. 
These are depicted in Fig.~\ref{fig:OverviewSystem}.
To train these LSTM-seq2seq models, we use 
the \texttt{RMSProp} optimizer and $\ell_2$-norm kernel regularizer of $1e-4$ to minimize the mean squared reconstruction error.
The output of LSTM-decoder is dropped out with a drop-rate $0.3$, and then passes through a fully-connected layer with the linear activation function to generate a reconstruction sequence.

\subsubsection{Anomaly Score}
We assume that reconstruction errors follow the Gaussian distribution
$\mathcal N(\mathbf{\mu}, \mathbf{\Sigma})$, where $\mathbf\mu$ and $\mathbf{\Sigma}$ are the mean and covariance matrix of reconstruction errors of normal data samples.
We use \textit{logarithmic probability densities (logPD)} of the reconstruction errors as \textit{anomaly scores}, as is similar to \cite{singh2017anomaly, Malhotra_LSTM_encDec_ICMLWrsh2016, Mao2020adaptive}.
We then use the minimum value of the logPD on the normal dataset (i.e., the training set) as the threshold for detecting outliers. 

We consider a detection as confident if the input sequence being detected satisfies one of these two conditions:
(i) at least one data point has a logPD of less than certain times (e.g., 2x) of the threshold (note logPD is negative);
(ii) the number of anomalous points is higher than a certain percentage (e.g., 5\%) of the sequence size.

\subsection{Adaptive Model Selection Scheme}
\label{subsec:DynamicModelSelectionScheme}

We propose an adaptive model selection scheme to select the most suitable AD model based on the contextual information of input data, so that each data sample will be directly fed to its best-suited model.
This is in contrast to traditional approaches where input data either (i) always go to one same model regardless of the hardness of detection \cite{Chen_SEC2017}, or (ii) are successively offloaded to higher layers until meeting a desired accuracy or confidence \cite{Teerapittayanon_ICDCS2017}. 

\begin{figure}[ht]
    \centering
    \includegraphics[width=0.85\linewidth]{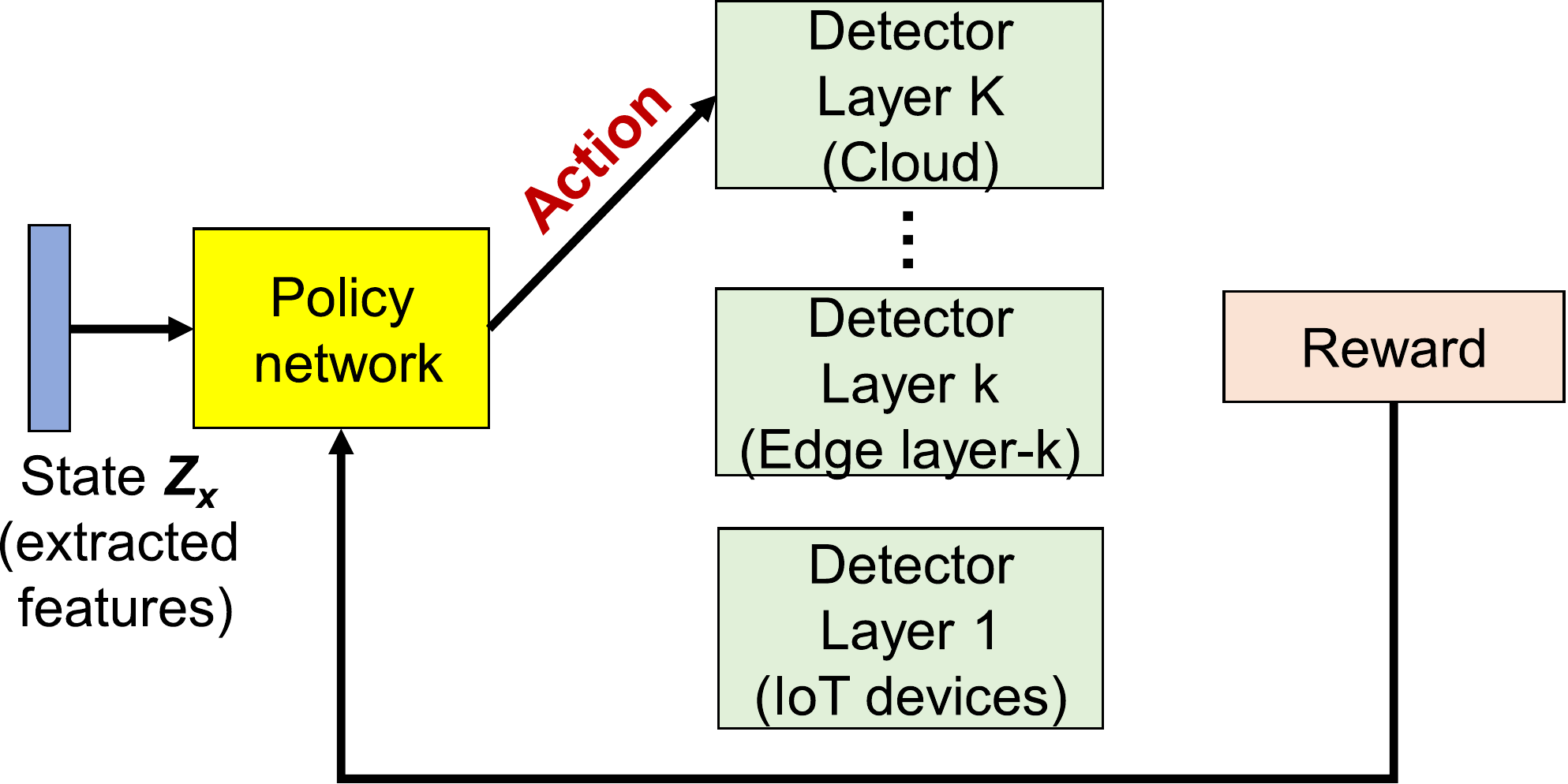}
    \caption{Adaptive model selection with a policy network.}
    \label{fig:DecisionMakingModule}
\end{figure}

Our proposed adaptive scheme is a reinforcement learning algorithm that adapts its model selection strategy to maximize the expected reward of a model to be selected.
We frame the learning problem as a \textit{contextual bandit problem} \cite{sutton2000policy,williams1992REINFORCE} 
and use a {\em policy gradient method} to solve it.
See Fig.~\ref{fig:DecisionMakingModule}.
Formally, given the contextual information $\mathbf{z_x}$ of an input data $\mathbf{x}$,
and $K$ trained AD models deployed at the $K$ layers of an HEC system,
we build a policy network that takes $\mathbf{z_x}$ as the input state and outputs a policy of selecting which model (or equivalently which layer of HEC) to do anomaly detection, in the form of a categorical distribution
$\pi_{\theta} (\mathbf{a}|\mathbf{z_x}) = \prod_{k=1}^{K} s_k^{a_k} ,$
where $\mathbf{a}$ is the actions encoded as a one-hot vector that defines which model (or HEC layer) to perform the task,
$\mathbf{s}=(s_1, s_2,\cdots, s_K)=f_{\theta}(\mathbf{z_x})$, is a vector representing the likelihood of selecting each model $k$. 
We denote the selected action as $|\mathbf{a}|=k$ if $k=\arg \max_{k} (s_k)$.

The policy network $f_{\theta}(.)$ is designed as a neural networks with parameters $\theta$. To make the policy network small enough to run fast on IoT devices, we use the extracted features $\mathbf{z_x}$ instead of the raw input data $\mathbf{x}$, to represent the contextual information of input data. 
We train the policy network using policy gradient method \cite{williams1992REINFORCE,sutton2000policy} to find an optimal policy $\pi$ that maps an input state $\mathbf{z_x}$ to an action (which model or layer) to minimize the negative expected reward: 
\[
\min L(\theta) = -\mathop{\mathbb{E}}_{\mathbf{a} \sim \pi_{\theta}}[R(\mathbf{a},\mathbf{z_x})],
\]
where $R(\mathbf{a},\mathbf{z_x})$ is a reward function of action $\mathbf{a}$ given state $\mathbf{z_x}$.
To reduce the variance of reward value and increase the convergence rate, we utilize {\em reinforcement comparison} \cite{williams1992REINFORCE} with a baseline $R(\tilde{\mathbf{a}}, \mathbf{z_x})$. 
In order to encourage selecting an appropriate model that jointly increases accuracy and reduces the cost of offloading tasks further away from the edge, we propose a reward function as follows:
$
R(\mathbf{a}, \mathbf{z_x}) = \text{accuracy}(\mathbf{x}) -  C(\mathbf{a}, \mathbf{x}).
$
We define the cost function $C(\mathbf{a}, \mathbf{x})$ as a function maps the end-to-end detection delay $t_\text{e2e}(\mathbf{x}, \mathbf{a})$ to an equivalent accuracy in scale $[0,1]$ with intuition that a higher delay will result in a greater reduction of accuracy:
\begin{align}
\label{eq:costFunction}
C(\mathbf{a}, \mathbf{x}) = \frac{\alpha \cdot t_\text{e2e}(\mathbf{x}, \mathbf{a}) }{1+ \alpha \cdot t_\text{e2e}(\mathbf{x}, \mathbf{a})},
\end{align}
where $\alpha$ is a tunable parameter. Further details are available in \cite{Mao2020adaptive}.


\section{Implementation and Experiment Setup}
\label{sec:ExpSetup}
\subsection{Dataset}
We evaluate our proposed approach with two public datasets. The data is standardized to zero mean and unit variance for all of the above training tasks and datasets.

\textbf{Univariate dataset.} We use a dataset on power consumption,\footnote{\url{http://www.cs.ucr.edu/~eamonn/discords/}}
which is used in \cite{Malhotra_LSTM_encDec_ICMLWrsh2016,singh2017anomaly, Mao2020adaptive}. For training details, please refer to our previous work \cite{Mao2020adaptive}. 

\textbf{Multivariate dataset.} We use  MHEALTH\footnote{http://archive.ics.uci.edu/ml/datasets/mhealth+dataset} which consists of 12 human activities of 10 different people, and each person wore two motion sensors: one on left-ankle and the other on right-wrist. Each motion sensor contains a 3-axis accelerator, a 3-axis gyroscope, and a 3-axis magnetometer; hence the input data has 18 dimensions. The sampling rate of these sensors is 50\,Hz.
We use a window sequence of 128 time-steps ($\sim$2.56 second) with a step-size of 64. 
Adopting the common practice, we choose the dominant human activity (e.g., walking) as normal and treat the other activities as anomalous.
For the AD task, we select 70\% of normal samples of all the subjects (people) as the training set; and the rest 30\% of normal samples plus 5\% of each of the other activities as the test set.
To train the policy network, we select 30\% of normal samples and 5\% of each of the other activities as the training set, and the whole dataset as the test set.

\subsection{Implementation}
\label{subsec:ExperimentSetup}

We use Tensorflow and Keras to implement the AD models (i.e., three AE models and three LSTM-seq2seq models as shown in Fig.~\ref{fig:OverviewSystem}) and the policy network model.
Note that, since the edge server and cloud server are empowered with GPU, we implement the LSTM-seq2seq-Edge and BiLSTM-seq2seq-Cloud models based on CuDNNLSTM units to accelerate training and inference time.
Before deploying the LSTM-seq2seq-IoT and LSTM-seq2seq-Edge models on Raspberry Pi3 and Jetson-TX2, we compress them by (i) removing the trainable nodes from the graph, and convert variables into constants; (ii) quantizing the model parameters from floating-point 32-bit (FP32) to FP16.
We observe no performance decrease of these compressed AD models running on Raspberry Pi3 and Jetson-TX2.

The policy network requires low complexity and needs to run fast on IoT devices, 
so the state input to the policy network needs to be small but still well represent the whole sequence of input data.
For the univariate data, we define the contextual state as an extracted feature vector which includes min, max, mean, and standard deviation of each day's sensor data.
For the multivariate data, we use the encoded states of the LSTM-encoder to represent the input state for the policy network.
Subsequently, we build the policy network as a single hidden neural network with 100 hidden units and an output layer with 3 units. 
We empirically select $\alpha=0.0005$ for the univariate dataset, and $\alpha=0.00035$ for the multivariate dataset to calculate the cost of executing detection as given by \eqref{eq:costFunction}.

\subsection{Software Architecture and Experiment Setup}
\begin{figure}
    \centering
    \subfloat[GUI: user can select dataset and evaluation scheme, and tune parameters.]{
        \includegraphics[width=1.0\linewidth]{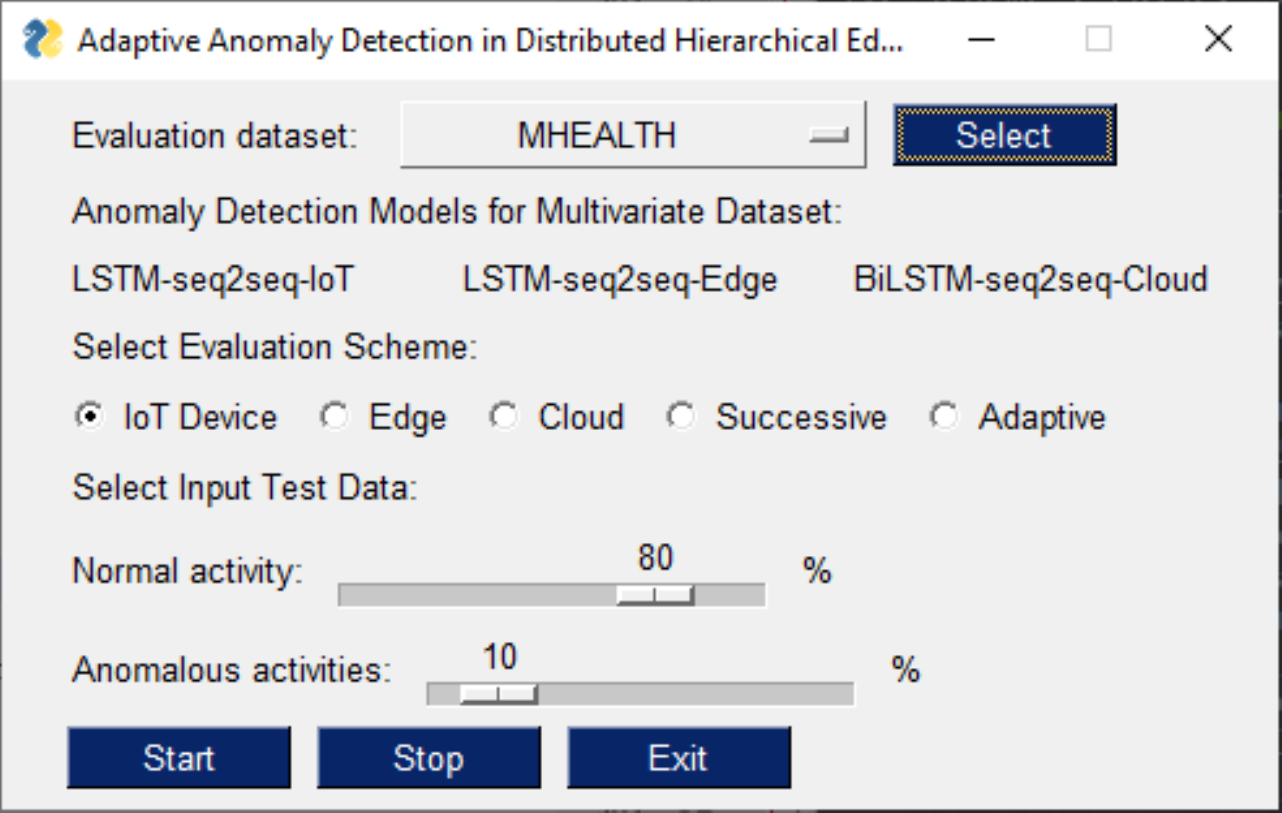}%
        \label{fig:DemoInterface}
    }

    \subfloat[Result panel: raw sensory signals, AD performance and associated actions.]{
        \includegraphics[width=1.0\linewidth]{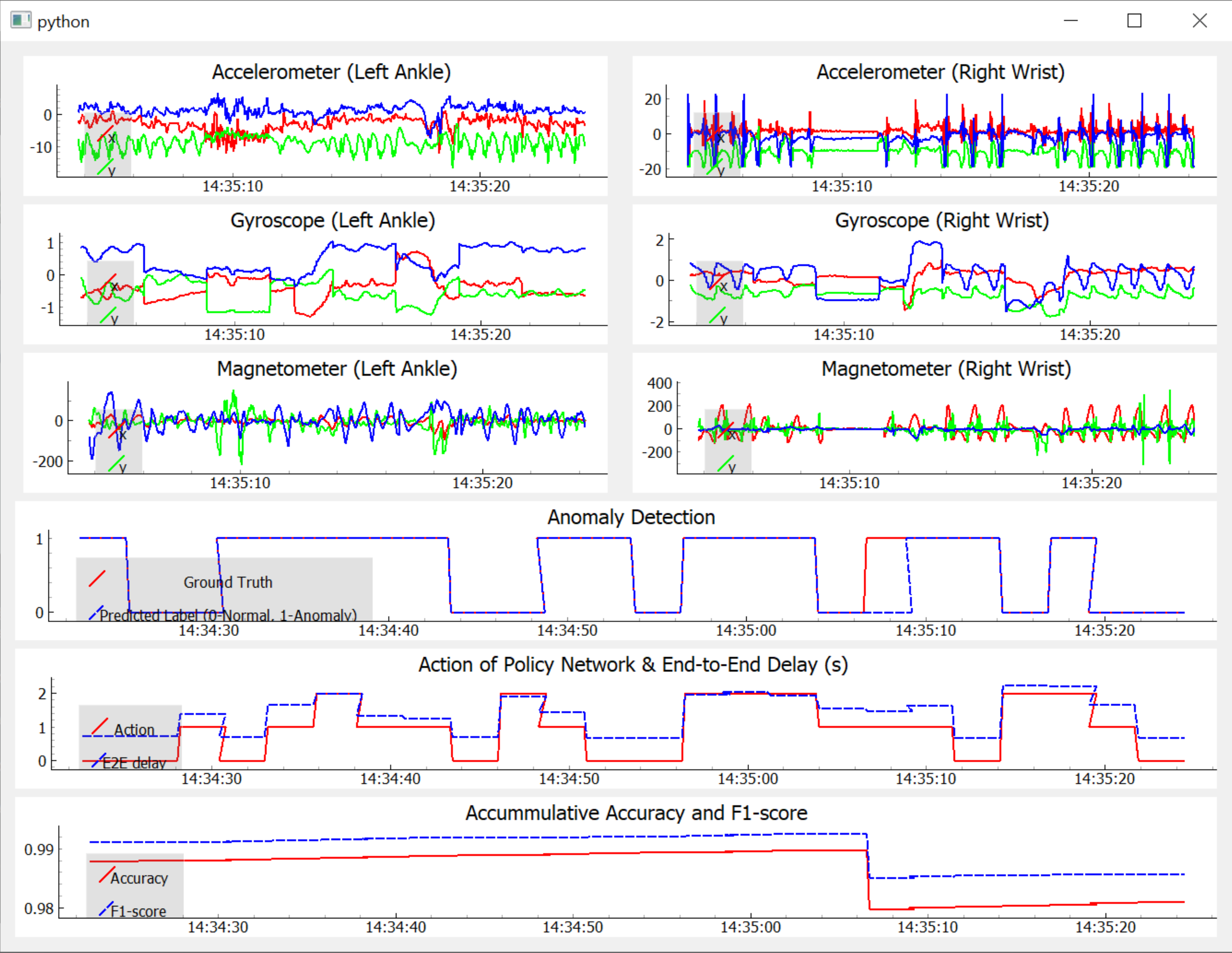}%
        \label{fig:DemoGraph}
    }
\caption{Demo: GUI and results: multivariate data.}
\label{fig:DemoGUI}
\end{figure}

\begin{figure*}[t]
	\centering
	\includegraphics[width=.7\linewidth]{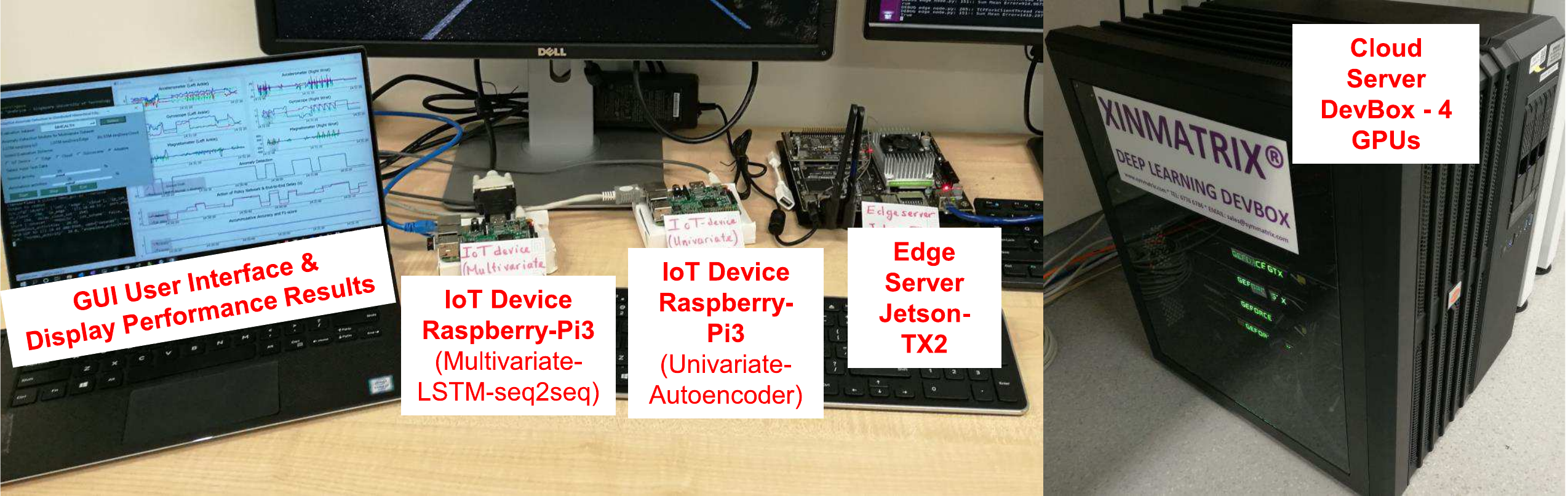}%
	\caption{Hardware setup of our HEC testbed.}
	\label{fig:HardwareTestbed}
\end{figure*}

The software architecture of our demo is shown in Fig.~\ref{fig:SoftwareArchitecture}. It consists of a GUI, the adaptive model selection scheme based on the policy network, and the three AD models at the three layers of HEC.
The GUI allows a user to select which dataset and model selection scheme to use, as well as to tune parameters, as shown in Fig.~\ref{fig:DemoInterface}, and displays the sensory raw signals and performance results, as shown in Fig.~\ref{fig:DemoGraph}.
All the communication services use keep-alive TCP sockets 
to reduce the overhead of connection establishment.
Network latency as shown in Fig.~\ref{fig:OverviewSystem} is configured by using Linux traffic control tool, \texttt{tc}, to emulate the WAN connections in HEC. The hardware setup for the HEC testbed is shown in Fig.~\ref{fig:HardwareTestbed}.

\textbf{User Actions:}
As shown in Fig.~\ref{fig:DemoInterface}, we allow users (e.g., ICDCS demo session participants) to evaluate the HEC testbed performance with (i) either univariate or multivariate datasets, (ii) different fractions of normal and abnormal data in the datasets to use, and (iii) different schemes under evaluation:
(1) always detects anomaly at \textit{IoT Device},
(2) always offloads detection tasks to \textit{Edge} server,
(3) always offloads to \textit{Cloud},
(4) \textit{Successive}, i.e., executes at IoT devices first and then offloads to higher layers successively until reaching a confident output or the cloud, and
(5) \textit{Adaptive} which is our proposed adaptive model selection scheme.
After the user clicks ``Start'', our result panel as in Fig.~\ref{fig:DemoGraph} will show the continuously updated raw sensory data (accelerometer, gyroscope, magnetometer), anomaly detection outcome (0 or 1) vs. ground truth, detection delay vs. the actions determined by our policy network, and the accumulative accuracy and F1-score.

\section{Experiment Results}
\label{subsec:ExperimentResults}
\textbf{Comparison among AD models:}
Table~\ref{tab:threeAnomalyDetectors} compares the performance and complexity among the three models we use. For both univariate and multivariate data, the complexity of AD models increases from IoT to cloud, as indicated by ``\# of Parameters'' (weights and biases) which reflects the approximate {\em memory footprint} of the models.
Along with this, the F1-score and accuracy increase as well; for example, AE-Cloud are 95.5\% and 20\% higher than AE-IoT, and BiLSTM-seq2seq-Cloud are 15\% and 14.74\% higher than LSTM-seq2seq-IoT on these two metrics.
On the other hand, the execution time (for running the detection algorithms) decreases from IoT to cloud, as indicated in the last row of Table~\ref{tab:threeAnomalyDetectors}, which is measured on the actual machines of our HEC testbed and averaged over five runs. This is due to the different computation capacity (whereas communication capacity is taken into account by our end-to-end delay shown later). One more observation is that LSTM-seq2seq models, which handle multivariate datasets, take much longer time to run (up to 591 ms) than AE models, which handle univariate datasets (up to 12.4 ms).

\begin{table}[t]
	\caption{Comparison among AD models.}
	\label{tab:threeAnomalyDetectors}
	\centering
	\scriptsize
    \begin{tabular}{ l@{\hspace{0.8em}} l@{\hspace{0.8em}} l@{\hspace{0.8em}} l@{\hspace{0.8em}}  l@{\hspace{0.8em}} l@{\hspace{0.8em}} l@{\hspace{0.8em}} }
		\toprule
		\textbf{Dataset/Model} & \multicolumn{3}{l}{\textbf{Univariate/Autoencoder}} & \multicolumn{3}{l}{\textbf{Multivariate/LSTM-seq2seq} }\\
		\midrule
		\textbf{Layer} & \textbf{IoT} & \textbf{Edge} & \textbf{Cloud} & \textbf{IoT} & \textbf{Edge} & \textbf{Cloud} \\
		\midrule
		\textbf{\#Parameters}        & 271,017 & 949,468 & 1,085,077  & 28,518 & 97,818 & 1,028,018 \\
		\textbf{Accuracy(\%)}        & 78.09   & 93.33 & 98.09         & 82.63 & 94.21 & 97.37 \\
		\textbf{F1-score}            & 0.465   & 0.741 & 0.909         & 0.852 & 0.955 & 0.980 \\
		\textbf{Exec time (ms)}    & 12.4    & 7.4  & 4.5          & 591.0  & 417.3 & 232.3 \\ 
		\bottomrule
	\end{tabular}
\end{table}

\textbf{Comparison among model selection schemes:}
We can see in Table \ref{tab:ExperimentResultDynamicScheme} that the IoT Device scheme achieves the lowest detection delay but also the poorest accuracy and F1-score among all the evaluated schemes.
On the other extreme, the Cloud scheme yields the best accuracy and F1-score but incurs the highest detection delay (end-to-end).
The Successive scheme leverages distributed anomaly detectors in HEC and thus significantly reduces the average detection delay as compared to the Edge and Cloud schemes.
However, its accuracy and F1-score are outperformed by the Edge scheme.
In contrast, our proposed adaptive scheme 
not only achieves lower detection delay but its F1-score and accuracy also consistently outperform those of IoT Device, Edge, and Successive schemes.
Even though the F1-score and accuracy of our proposed scheme are marginally lower than those of the Cloud scheme (e.g., 0.82\% and 0.4\% for the multivariate dataset), 
our scheme reduces the end-to-end detection delay by a substantial 71.4\% and 7.84\% for the univariate and multivariate datasets, respectively. 
In summary, and as also indicated in the last column ``Reward'', which is a convex combination of both accuracy and delay, our proposed adaptive scheme strikes the best tradeoff between accuracy and detection delay.

\begin{table}[t]
    \caption{Comparison among AD model detection schemes.}
    \label{tab:ExperimentResultDynamicScheme}
    \centering
    \footnotesize
    \begin{tabular}{l@{\hspace{0.8em}} l@{\hspace{1em}} c@{\hspace{1em}}   c@{\hspace{1.em}}  c@{\hspace{1.em}}  c@{\hspace{1.em}} }
        \toprule
         \textbf{Dataset} & \textbf{Scheme}  &  \textbf{F1} & \textbf{Accuracy(\%)} & \textbf{Delay(ms)} & \textbf{Reward}\\%
        \midrule%
        \multirow{5}{*}{\STAB{\rotatebox[origin=c]{90}{Univariate}} } &        \textbf{IoT Device}    & 0.465 & 93.68 & 12.4 & 48.39 \\%
        & \textbf{Edge}   & 0.800 & 98.63 & 257.43 & 45.36 \\%
        & \textbf{Cloud}  & 0.909 & 99.46 & 504.50 & 41.24\\%
        & \textbf{Successive} & 0.769 &  98.35 & 105.27 &   N/A   \\%
        & \textbf{Our Method}  & 0.870   & 99.17 & 144.50 & {\bf 49.52} \\%
        \midrule
        \multirow{5}{*}{\STAB{\rotatebox[origin=c]{90}{Multivariate}} } &\textbf{IoT Device}    & 0.848 & 93.19 & 591.0 & 389.85\\%
        & \textbf{Edge}   & 0.951 & 97.59 & 667.30 & 403.77 \\%
        & \textbf{Cloud}  & 0.980 & 99.00 & 732.30 & 404.12\\%
        & \textbf{Successive} & 0.911 &  95.79  & 626.16 &   N/A  \\%
        & \textbf{Our Method}  & 0.972   & 98.60 & 674.87  & {\bf 408.06}\\%
        \bottomrule
    \end{tabular}
\end{table}


\section{Conclusion}
\label{sec:conclusions}

In this paper, we identify three issues in existing IoT anomaly detection (AD) approaches, 
and propose an adaptive AD approach for IoT data in HEC for both univariate and multivariate datasets.
We construct multiple distributed AD models based on autoencoder and LSTM with increasing complexity, and associate them with the HEC layers from bottom to top. 
Next, it uses a reinforcement learning based adaptive scheme to select the best-suited model based on the contextual information of input data. The scheme consists of a policy network as the solution to a contextual bandit problem, characterized by a single-step MDP.
We build an HEC testbed and conduct our demo experiments using two real-world IoT datasets. By comparing with other baseline schemes, we show that our proposed scheme strikes the best performance tradeoff between detection accuracy and detection delay, which is a typical dilemma.


\bibliographystyle{IEEEtran}
\small{
\bibliography{references}
}

\end{document}